\documentclass[letterpaper]{article} 
\usepackage{aaai2027}  
\usepackage[hyphens]{url}  
\usepackage{graphicx} 
\usepackage{amsmath}
\usepackage{amsfonts}
\usepackage{natbib}  
\usepackage{caption} 
\usepackage{algorithm}
\usepackage{algorithmic}
\usepackage{url}
\usepackage{newfloat}
\usepackage{listings}
\DeclareCaptionStyle{ruled}{labelfont=normalfont,labelsep=colon,strut=off} 
\floatstyle{ruled}
\newfloat{listing}{tb}{lst}{}
\floatname{listing}{Listing}

\usepackage{booktabs}

\title{ReactHuman: A Physics-Grounded Benchmark for\\ Human-Like Reactive Decision-Making in Embodied Multimodal LLMs}
\author {
    Yizhan Li\textsuperscript{\rm 1,\rm 2},
    Jianxin You\textsuperscript{\rm 1,\rm 2},
    Mengyang Xiong\textsuperscript{\rm 3,\rm 2},
    Yinhuan Chen\textsuperscript{\rm 3},
    Zicheng Zhao\textsuperscript{\rm 4},
    Dekun Wu\textsuperscript{\rm 1,\rm 2},
    Dongqing Zhang\textsuperscript{\rm 5},
    Bang Liu\textsuperscript{\rm 1,\rm 2}\corresponding
}
\affiliations {
    \textsuperscript{\rm 1}Université de Montréal\\
    \textsuperscript{\rm 2}Mila -- Quebec AI Institute\\
    \textsuperscript{\rm 3}McGill University\\
    \textsuperscript{\rm 4}McMaster University\\
    \textsuperscript{\rm 5}Meta Platforms\\
    \{yizhan.li, jianxin.you, dekun.wu, bang.liu\}@umontreal.ca,
    \{mengyang.xiong, yinhuan.chen\}@mail.mcgill.ca,
    zhaoz149@mcmaster.ca,
    zhangdq20@gmail.com
}

\begin{document}

\maketitle

\begin{abstract}
Reacting to sudden physical hazards (catching a slipping plate, dodging a falling knife) is both a meaningful test of embodied intelligence and a hard requirement for deploying multimodal large language models (MLLMs) as the
decision core of household robots.
Existing evaluations, however, probe intuitive physics passively through
question answering over videos, or target deliberate, long-horizon tasks such
as navigation and rearrangement; none measure whether a model can turn
physical understanding into immediate, safety-critical action.
We introduce ReactHuman, the first physics-grounded benchmark for human-like
reactive decision-making, in which the evaluated MLLM acts as the brain of a
simulated humanoid facing sudden household hazards; it spans 17 event families
and over 1{,}000 bit-for-bit reproducible scenes with exact, annotation-free
ground truth derived from 240\,Hz rigid-body simulation, including adversarial
objects whose appearance contradicts their physics (a foam anvil, a steel
apple).
We further design a five-metric suite that scores each reaction along three
axes: reasonable, safe, and physically grounded. We physically execute
every committed plan so that decisions have observable consequences. With this
harness we evaluate seven representative MLLMs.
Results show that reactive safety is far from solved: models mishandle roughly
one hazard in three, act from fixed dispositions rather than the observed
scene, trust appearance over motion, and miss interception points at meter
scale even when the chosen action is correct; none of these failures
shrink with model scale.
ReactHuman thus offers both a fine-grained diagnosis and a scalable training
signal toward physically grounded, safety-aware embodied agents. The benchmark can be found here: \url{https://huggingface.co/datasets/Alan123/reacthuman-benchmark-scaled}
\end{abstract}


\section{Introduction}
Humans react to sudden physical hazards in a fraction of a second: we catch a
slipping plate, dodge a falling knife, and brace against a toppling wardrobe.
These reactions fuse \emph{semantic} judgment (what is this object, and is it
safe to touch?) with \emph{intuitive physics} (where and when will it arrive?).
As multimodal large language models (MLLMs) are placed at the cognitive core of
household robots and embodied assistants \citep{ahn2022saycan,driess2023palme},
the same competence becomes a deployment requirement: we want an agent that
reacts \emph{like a competent person}, whose reaction is at once
\emph{safe} (it avoids harm), \emph{reasonable} (it is the action a sensible
human would choose and defend), and \emph{physically grounded} (the hand
actually meets the object in time). An agent that elects to catch a falling
chef's knife, or that decides correctly yet grasps half a meter wide, has
failed in a way no captioning or VQA benchmark can reveal.

Existing evaluations do not measure this competence. Physical-reasoning
benchmarks probe intuitive physics \emph{passively}, through plausibility
judgments or question answering over videos
\citep{bear2021physion,yi2020clevrer,zheng2024contphy,chow2025physbench},
while embodied-AI benchmarks emphasize deliberate, long-horizon tasks such as
navigation and rearrangement \citep{szot2021habitat,li2023behavior}. Neither
reveals whether a model can convert physical understanding into an
\emph{immediate, safety-critical action} nor diagnose \emph{why} it fails
when it does: a model may misread the hazard, choose an unsafe action despite
understanding it, or choose correctly yet mispredict the interception point by a meter.

We introduce \textbf{ReactHuman}, a benchmark that evaluates whether MLLMs
react to sudden hazards \emph{like a competent human} by making the MLLM the decision-making brain of a simulated humanoid that must actually carry the reaction out. We check if they act safely, reasonably, and with physically grounded movement in physically simulated indoor scenes. ReactHuman adopts a
\emph{freeze-and-predict} protocol: a full-physics simulation of a sudden
household event runs to a critical decision moment and freezes; the model
observes the unfolding event and must commit to a structured reaction
plan: a walking command plus a hand-keyframe trajectory, from which the
executed action (\textsc{Catch}, \textsc{Dodge}, or \textsc{No-Action}) is
derived. Because inference happens while simulated time is
paused, models are compared on decision quality alone, uncontaminated by API
latency. Because every
scene is generated deterministically from a seed and labeled from simulator
state, all scores are computed against exact physical ground truth. And
because the committed plan is then \emph{executed by a simulated humanoid}
driven by a pre-trained RL walking policy
\citep{rudin2022learning,unitree2024rlgym} in the same physics engine, every decision has an observable physical
outcome (the catch connects or the toppling shelf strikes the agent) rather
than remaining a multiple-choice answer, which is what separates ReactHuman
from VQA-style physical-reasoning evaluation.

ReactHuman is built for \emph{diagnosis}, not a single leaderboard number.
Its \textbf{17 event families} (Figure~\ref{fig:taxonomy}) are chosen so that
different families stress different links in the perception--reasoning--action
chain: tabletop drops make the trajectory trivial and isolate semantic choice;
pendulums and bouncing balls make the semantic choice easy and isolate
interception; chain reactions require causal propagation before any kinematics;
\emph{adversarial objects} (a foam anvil, a steel apple, a lead-core tennis
ball) decouple appearance from physics, so that the correct action is
recoverable only from observed motion. Complementing the taxonomy, a
\textbf{five-metric suite} instruments the three axes of a human-like reaction:
semantic action accuracy and action--intent alignment (is the reaction
\emph{reasonable}?), safety validity (is it \emph{safe}?), and physical
endpoint distance and hand-distance evolution (is it \emph{physically
grounded}?).

Our contributions are:
\begin{itemize}
\item \textbf{Benchmark.} A physics-grounded, freeze-and-predict evaluation of
human-like reactive decisions along three axes: safety, reasonability, and
physical grounding. The evaluation covers 17 sudden-event families, over 1000 reproducible
scenes, three synchronized camera views, and exact simulator-derived ground
truth (action, 3D impact point, time-to-floor), including adversarial
appearance--physics probes, with every committed decision executed by a
simulated humanoid in the same physics engine.
\item \textbf{Generation pipeline.} A hybrid LLM-planned, seed-deterministic
pipeline in which a frozen LLM contributes semantic diversity (objects, rooms,
event routing from natural-language descriptions) while a procedural randomizer
owns every physical parameter. This makes scenes bit-for-bit reproducible and the
benchmark extensible past $10^4$ scenes without human annotation.
\item \textbf{Metric suite.} Five complementary metrics organized along the
three axes: reasonability, safety, and physical grounding. This separates an
unreasonable choice from an unsafe one from a kinematically ungrounded one, with
an explicit safety-rule system and structured-output protocol.
\item \textbf{Empirical study.} An evaluation of seven MLLMs across 306
scenes and all 17 families, revealing that safety failures concentrate
where evasion is required, that action choice follows fixed per-model
dispositions rather than the scene, that neither accuracy nor safety
improves with model scale, and that models never revise appearance-based
physics judgments from observed motion.
\end{itemize}

\section{Related Work}
\paragraph{Physical reasoning benchmarks.}
Physion and Physion++ \citep{bear2021physion,tung2023physion} test forward
prediction of physical outcomes; CLEVRER \citep{yi2020clevrer} and IntPhys
\citep{riochet2018intphys} probe causal and violation-of-expectation
reasoning; ContPhy \citep{zheng2024contphy} extends to deformables and fluids.
For modern MLLMs, PhysBench \citep{chow2025physbench} is the closest neighbor:
it measures physical-world understanding via multiple-choice VQA over a model
population that largely overlaps ours. Physics-IQ \citep{motamed2025physicsiq}
scores video-generation models for physical plausibility. ReactHuman differs on two axes simultaneously: the model's
output is an \emph{action with safety consequences} rather than an answer, and
ground truth is generated automatically and exactly by simulation, so the
quantitative track scales.

\paragraph{Embodied-AI and humanoid benchmarks.}
Habitat \citep{savva2019habitat,szot2021habitat}, AI2-THOR
\citep{kolve2017ai2thor}, ThreeDWorld \citep{gan2021threedworld}, ManiSkill~\citep{mu2021maniskillgeneralizablemanipulationskill}, ManiSkill2
\citep{gu2023maniskill2}, BEHAVIOR-1K \citep{li2023behavior}, and OpenEQA
\citep{majumdar2024openeqa} evaluate navigation, manipulation, rearrangement,
or situated QA on deliberate time scales; HumanoidBench
\citep{sferrazza2024humanoidbench} targets whole-body \emph{control}. None
evaluate immediate reactive safety. ReactHuman isolates the \emph{decision
layer} while retaining embodiment: following a brain--spine--body decoupling,
the evaluated MLLM acts as the brain of a simulated humanoid whose balance
and joint torques are delegated to a pre-trained whole-body controller
\citep{rudin2022learning,unitree2024rlgym}, so scores are attributable to reasoning rather than
policy quality, yet every decision is still physically executed in scene.

\paragraph{MLLMs as embodied decision modules vs.\ VLAs.}
One line of work deploys frozen MLLMs as zero-shot planners over discrete
skills \citep{ahn2022saycan,driess2023palme}; another fine-tunes
vision--language backbones end-to-end into vision--language--action (VLA)
policies emitting low-level control \citep{brohan2023rt2,kim2024openvla,
black2024pi0}. ReactHuman's harness spans the two: coupling a frozen MLLM to
a pre-trained whole-body controller and a simulated humanoid turns any
off-the-shelf MLLM into a zero-shot vision-to-action agent, without the
action fine-tuning that defines VLAs. We evaluate general-purpose MLLMs
because they are what is deployed today as the reasoning core of embodied
systems; native VLA policies can be dropped into the same scenes, ground
truth, and metrics unchanged---a comparison we leave to future work.

\paragraph{Procedural and LLM-assisted scene generation.}
Kubric \citep{greff2022kubric} established scalable simulator--renderer
tooling; Objaverse \citep{deitke2023objaverse} supplies large-scale assets;
recent systems use LLMs to compose environments and tasks
\citep{wang2024robogen,yang2024holodeck}. ReactHuman combines the two: an LLM
performs \emph{semantic} planning from natural language while a deterministic,
seeded randomizer retains exclusive control of \emph{physical} parameters,
preserving the reproducibility and label exactness that fully LLM-driven
generation lacks.
\section{The ReactHuman Benchmark}

\label{sec:benchmark}

\subsection{Freeze-and-Predict Protocol}
\label{sec:protocol}

Each episode places a virtual observer. A digital human standing in an indoor room is required to react to a sudden physical event at $t_0$: an object begins to
fall, slide, roll, topple, swing, or be struck toward the observer. The model
receives the \emph{observation window}: frames from $t_0$ to $t_0{+}\Delta$
(default $\Delta\!\approx\!0.6$\,s, before the outcome is visually resolved)
from up to three synchronized viewpoints (standing observer, close-up at the
event origin, overhead). Simulated time then freezes and the model must output
a structured reaction plan
\begin{equation}
\text{output} = \{\text{intent},\ \text{confidence},\
\text{walking\_cmd},\ \text{keyframes}\},
\end{equation}
where \texttt{intent} is
free-form reasoning, \texttt{confidence} $\in [0,1]$, \texttt{walking\_cmd} is
a base-velocity command, and \texttt{keyframes} $\{\mathbf{h}_1,\dots,
\mathbf{h}_n\}$, $\mathbf{h}_i\in\mathbb{R}^3$, is the predicted hand
trajectory. Freezing time makes the decision problem identical for every model
and removes inference latency as a confound; scoring is fully deterministic.

\begin{figure*}[t]
\centering
\setlength{\tabcolsep}{1.5pt}
\renewcommand{\arraystretch}{0.9}
\newcommand{\qv}[1]{\includegraphics[width=0.145\textwidth]{#1}}
\newcommand{\qrowA}[1]{%
\qv{Figures/qualitative_v3/col1_ceiling_drop/f#1.png} &
\qv{Figures/qualitative_v3/col2_stair/f#1.png} &
\qv{Figures/qualitative_v3/col3_ladder/f#1.png} &
\qv{Figures/qualitative_v3/col4_door/f#1.png} &
\qv{Figures/qualitative_v3/col5_pendulum/f#1.png} &
\qv{Figures/qualitative_v3/col6_drop/f#1.png}\\}
\newcommand{\qrowB}[1]{%
\qv{Figures/qualitative_v3/col7_furniture/f#1.png} &
\qv{Figures/qualitative_v3/col8_sliding/f#1.png} &
\qv{Figures/qualitative_v3/col9_shelf/f#1.png} &
\qv{Figures/qualitative_v3/col10_chain/f#1.png} &
\qv{Figures/qualitative_v3/col11_bouncing/f#1.png} &
\qv{Figures/qualitative_v3/col12_mirror/f#1.png}\\}
\resizebox{\textwidth}{!}{%
\begin{tabular}{@{}cccccc@{}}
{\footnotesize\textbf{Ceiling Drop}} & {\footnotesize\textbf{Stair Tumble}} &
{\footnotesize\textbf{Ladder Slip}} & {\footnotesize\textbf{Door Swing}} &
{\footnotesize\textbf{Pendulum}} & {\footnotesize\textbf{Object Drop}}\\
\qrowA{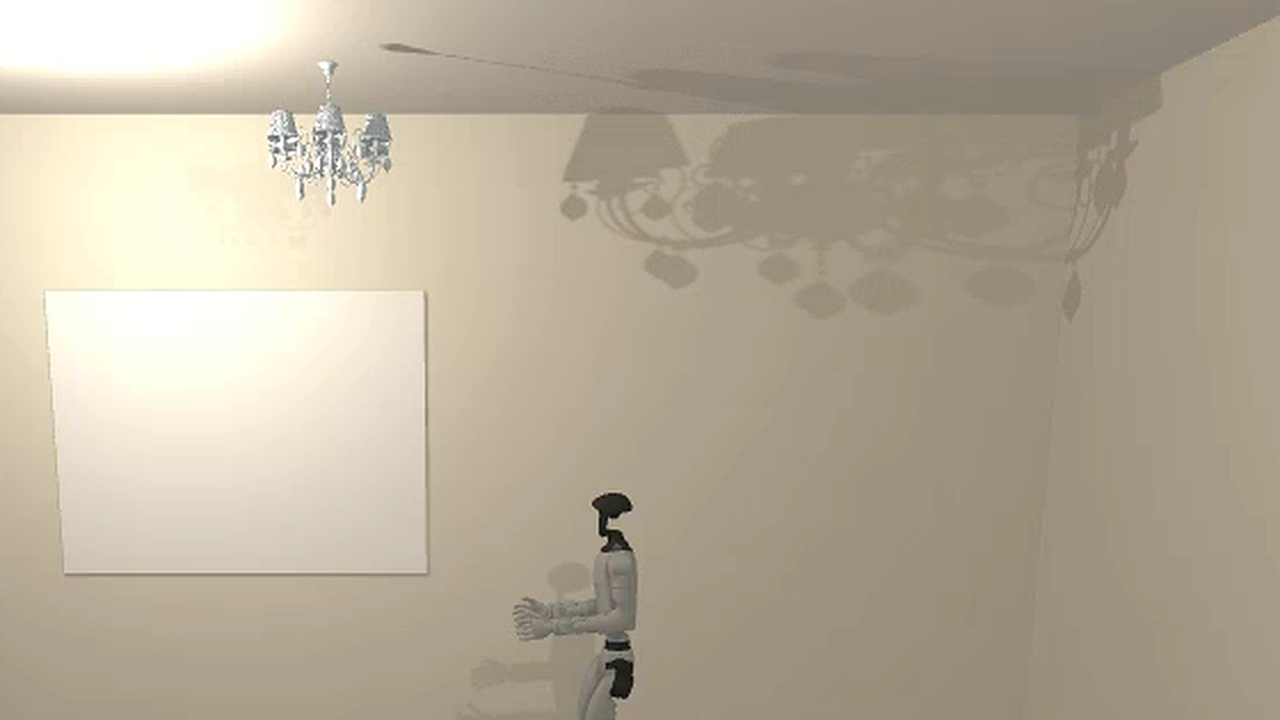}
\qrowA{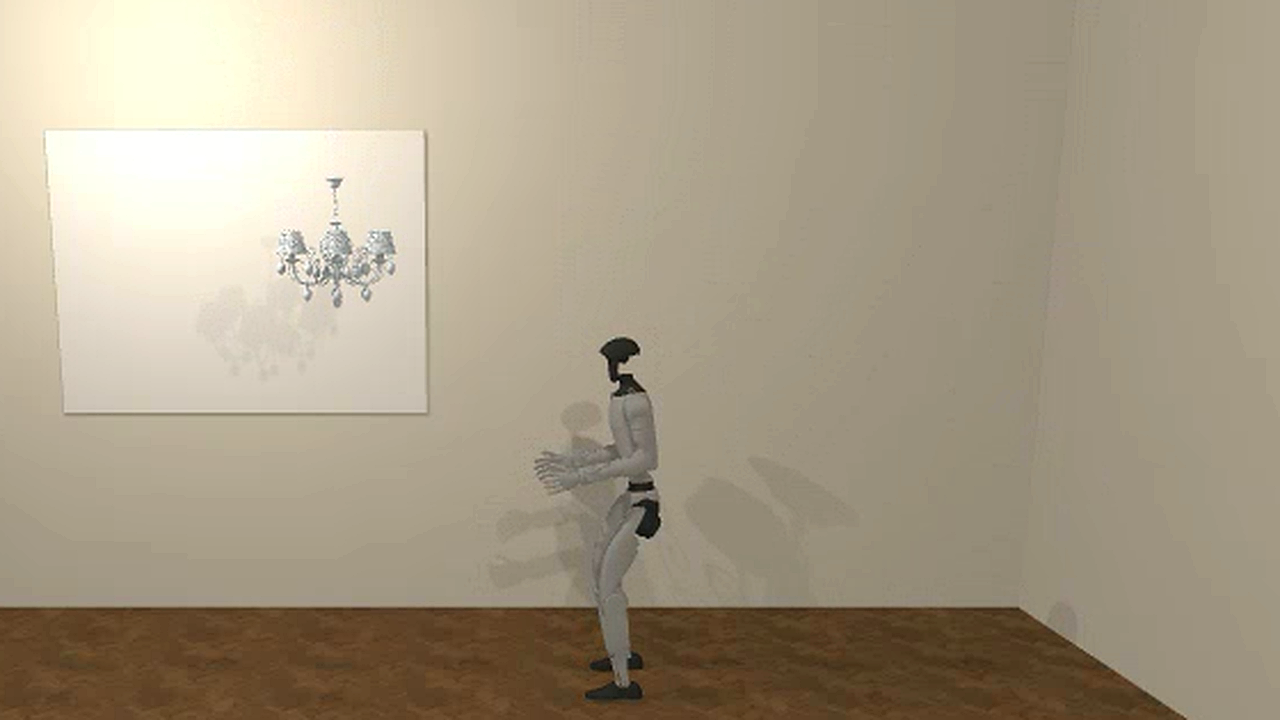}
\qrowA{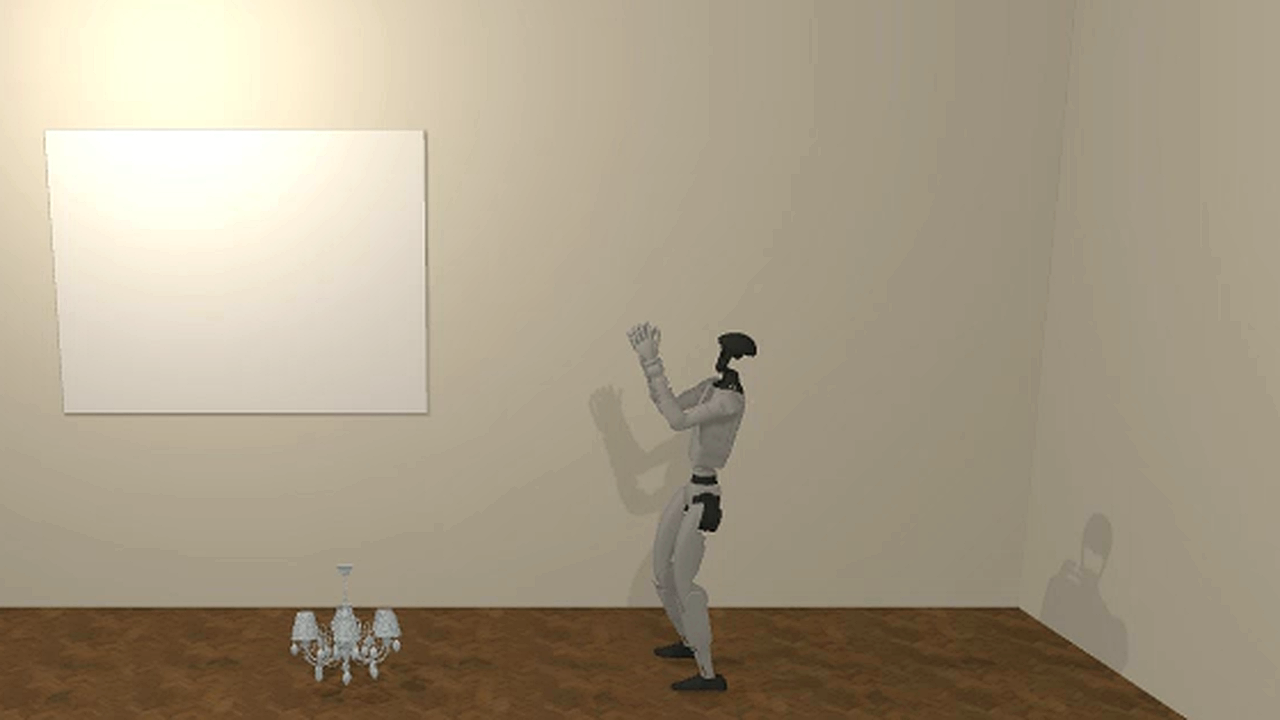}
\noalign{\vskip 4pt}
{\footnotesize\textbf{Furniture Tip}} & {\footnotesize\textbf{Sliding Object}} &
{\footnotesize\textbf{Shelf Slide}} & {\footnotesize\textbf{Chain Reaction}} &
{\footnotesize\textbf{Bouncing Object}} & {\footnotesize\textbf{Hanging Fall}}\\
\qrowB{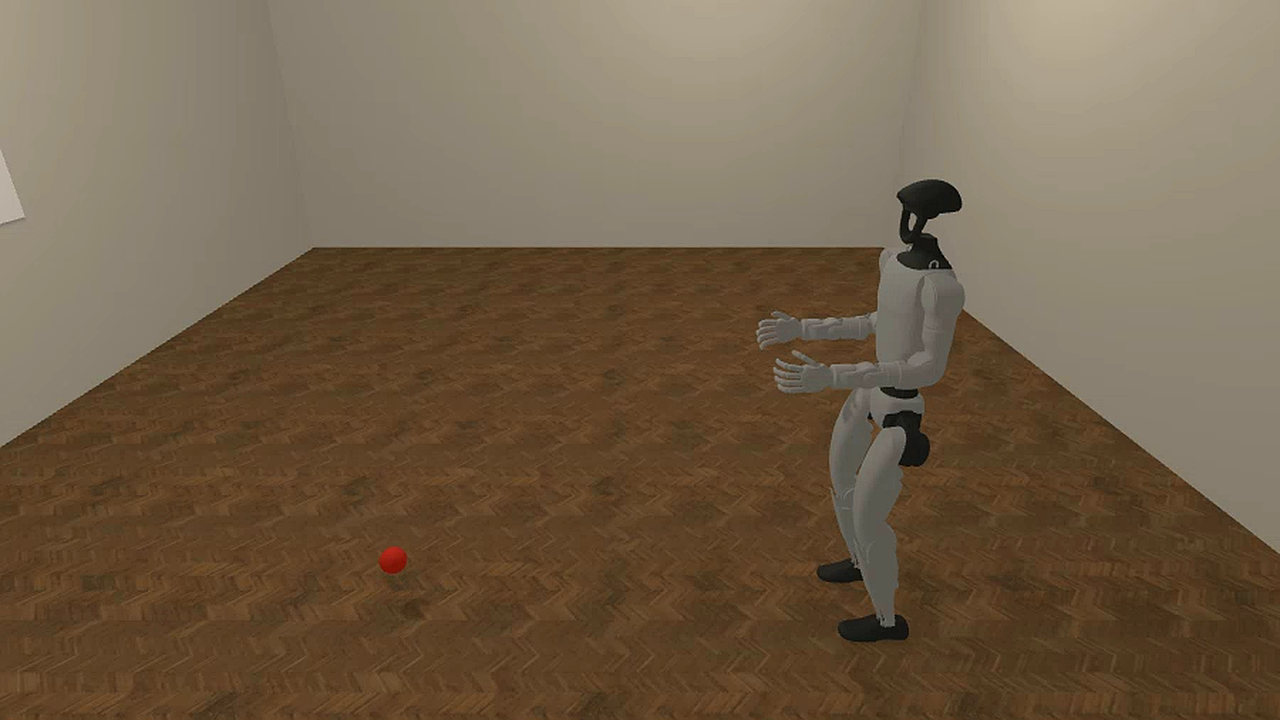}
\qrowB{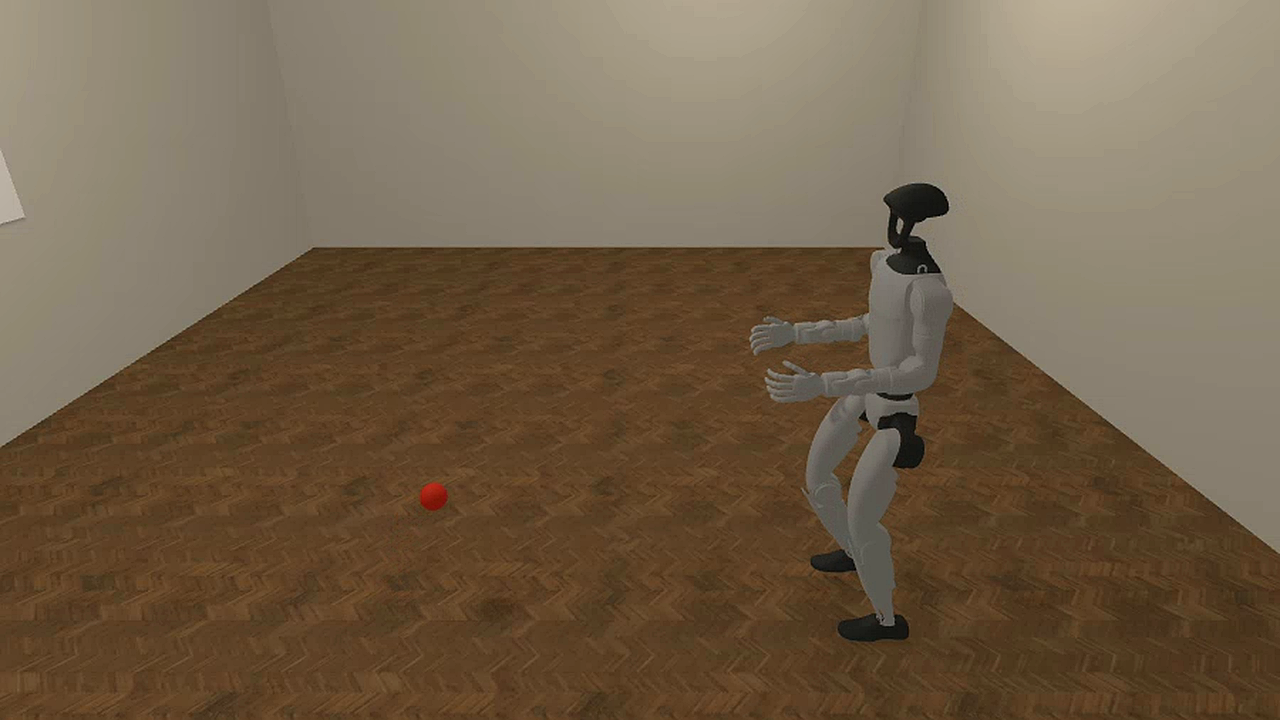}
\qrowB{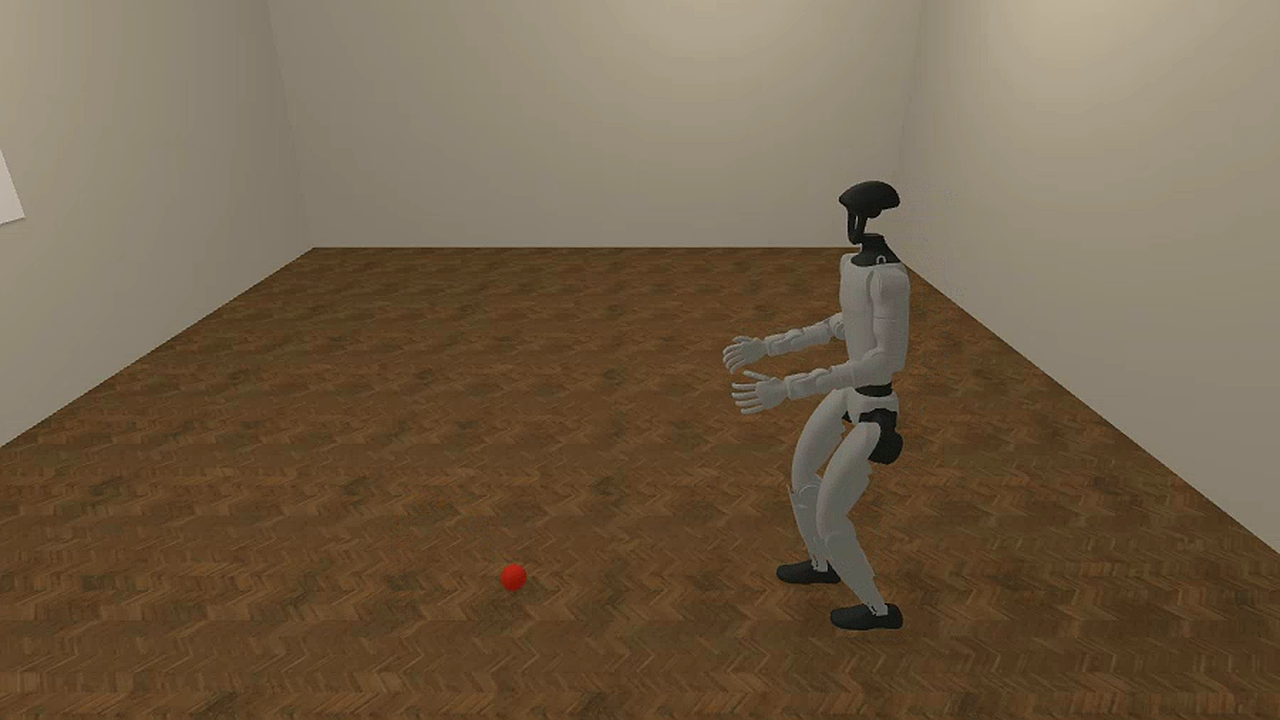}
\end{tabular}
}
\caption{\textbf{Twelve scenes from twelve event families.} Each column
is one scene read top to bottom (three frames, third-person view), with the
event family named above; all reactions shown are produced by Claude Opus
4.8. Every decision is physically executed, so each reaction ends in an
observable outcome.}
\label{fig:qualitative}
\end{figure*}

\paragraph{Humanoid execution.}
The reaction plan is not merely graded on paper: simulation then resumes and
the plan is \emph{executed by a simulated humanoid} (a Unitree G1) placed in
the same Genesis scene, following a brain--spine--body decoupling: the
evaluated MLLM is the \emph{brain}; a pre-trained whole-body controller (the
\emph{spine}) converts the walking command and hand keyframes into balance and
joint torques; the humanoid \emph{body} interacts with the falling object
under full rigid-body physics. Each episode therefore yields an execution
video in which the chosen reaction has observable physical
consequences. The catch connects or misses, the dodge clears the impact zone
or fails to. This distinguishes ReactHuman from passive VQA: the unit of
evaluation is an embodied interaction, not an answer. For metric
\emph{scoring} we use the simulator-derived ground truth
(\S\ref{sec:metrics}), which keeps scores deterministic and independent of
controller quality; execution provides outcome-level verification.
Representative rollouts are shown in Figure~\ref{fig:qualitative}.

\paragraph{Action space.} The model does not pick from a menu. It outputs
a motor plan, a base-velocity walking command plus hand keyframes, and a
deterministic rule classifies the plan into one of three primitives:
\textsc{Execute\_Catch} (walk toward the object and reach for it),
\textsc{Trigger\_Dodge} (move clear of its path), or \textsc{No\_Action}
(stay put, hands at rest). We score what the body would actually do, not
what the model claims, so a model cannot pass by simply saying the right
word. Ground-truth labels follow object properties and event kinematics,
not appearance: light, graspable, benign objects are labeled
\textsc{Catch}; sharp, hot, shattering, heavy, or fast objects are
labeled \textsc{Dodge}. \textsc{No\_Action} is correct only in the few
scenes where the event cannot reach the observer, and choosing it under
an active hazard counts as a safety violation. On the evaluated set
the labels split 141/156/9 across
\textsc{Catch}/\textsc{Dodge}/\textsc{No\_Action}, so majority-class
guessing (always \textsc{Dodge}) attains 51.0\%.

\subsection{A Taxonomy of 17 Sudden-Event Families}
\label{sec:taxonomy}
\begin{figure*}[t]
\centering
\setlength{\tabcolsep}{2pt}
\newcommand{\tx}[3]{\begin{minipage}[t]{0.148\textwidth}\centering
\includegraphics[width=\linewidth]{Figures/taxonomy_thumbs/#1.png}\par
\nointerlineskip\vspace{2pt}
{\fontsize{7}{7.6}\selectfont\textbf{#2}\\#3\par}
\end{minipage}}
\resizebox{\textwidth}{!}{%
\begin{tabular}{@{}cccccc@{}}
\tx{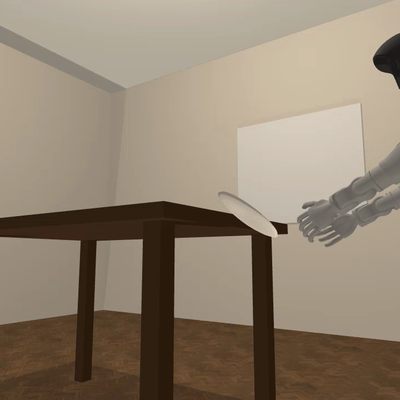}{object\_drop}{tips off a table edge} &
\tx{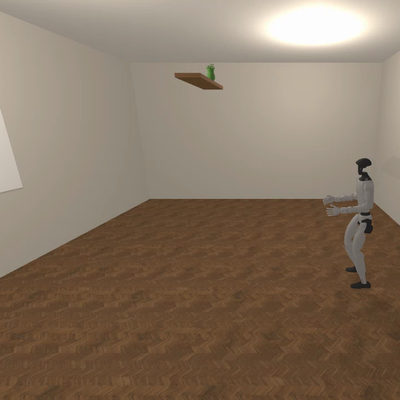}{shelf\_slide}{slides off a high shelf} &
\tx{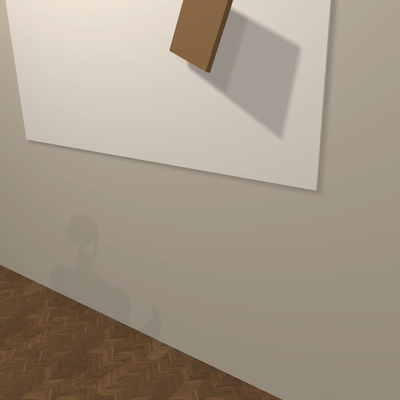}{hanging\_fall}{wall fixture detaches} &
\tx{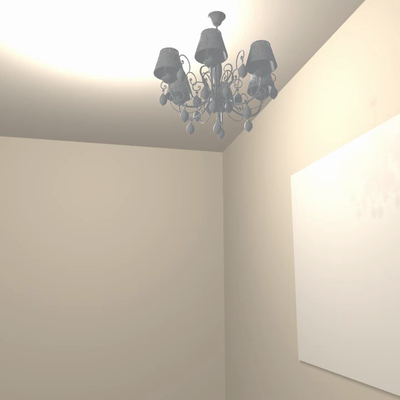}{ceiling\_drop}{falls from the ceiling} &
\tx{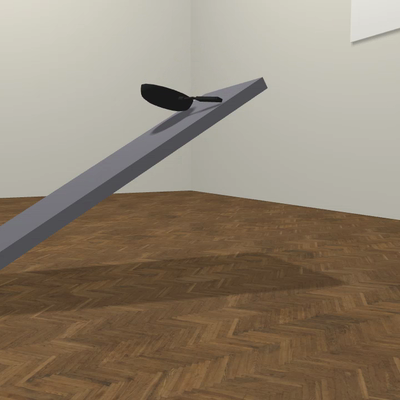}{sliding\_object}{slides down a ramp} &
\tx{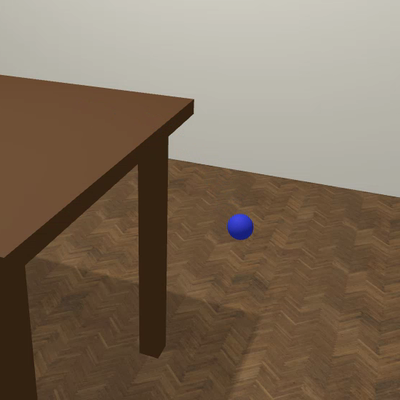}{rolling\_ball}{rolls off and across}\\[24pt]
\tx{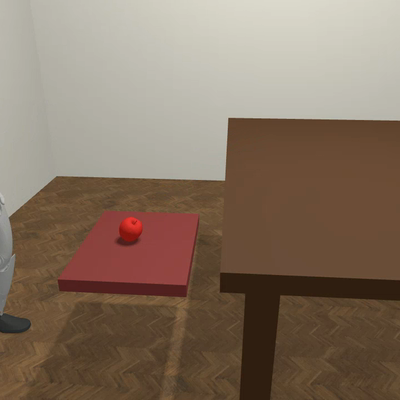}{surface\_cascade}{hops table, stool, floor} &
\tx{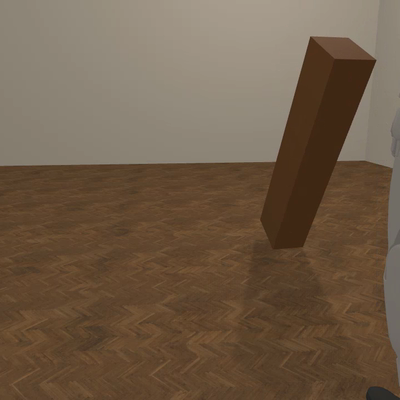}{furniture\_tip}{tall furniture tips over} &
\tx{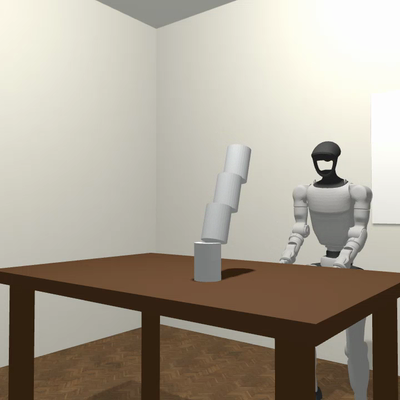}{stack\_collapse}{stacked items collapse} &
\tx{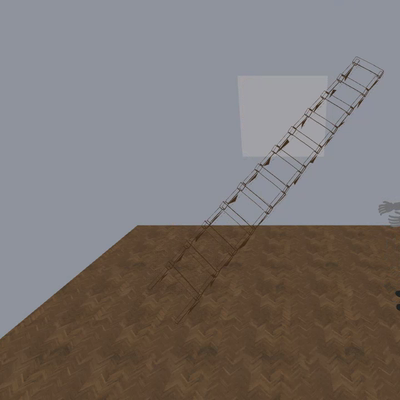}{ladder\_slip}{leaning ladder slips} &
\tx{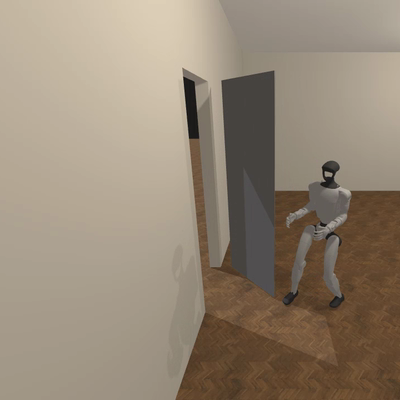}{door\_swing}{door swings violently} &
\tx{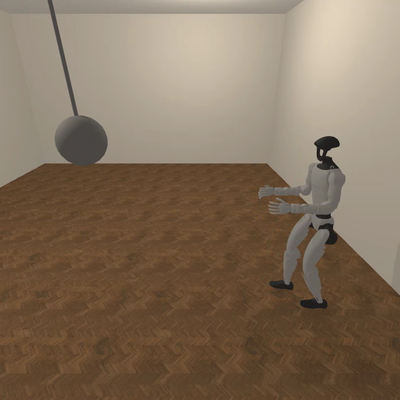}{pendulum\_swing}{suspended mass swings in}\\[24pt]
\multicolumn{6}{c}{%
\tx{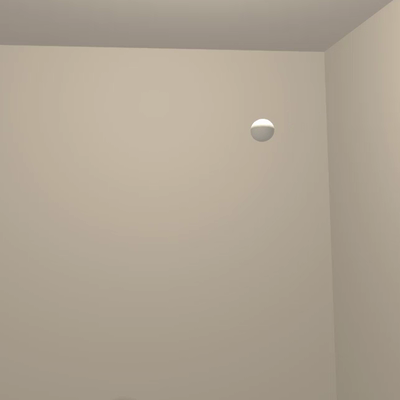}{thrown\_object}{thrown at the observer}\hspace{2pt}
\tx{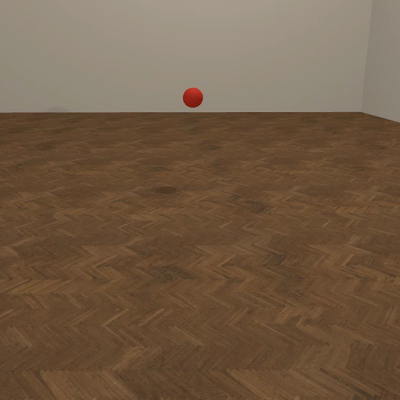}{bouncing\_object}{bounces toward observer}\hspace{2pt}
\tx{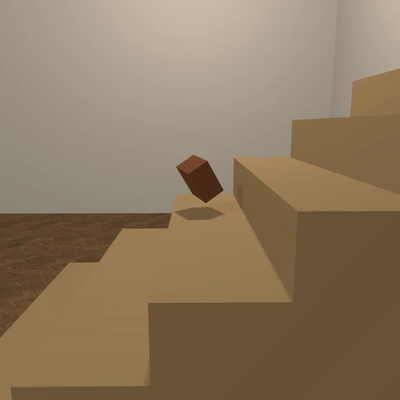}{stair\_tumble}{tumbles down the stairs}\hspace{2pt}
\tx{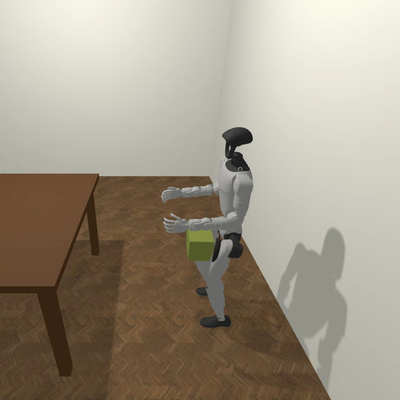}{chain\_reaction}{strike launches a second object}\hspace{2pt}
\tx{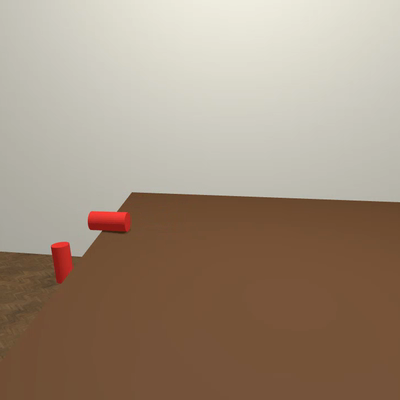}{multi\_object}{momentum runs down a row}}\\
\end{tabular}
}
\caption{\textbf{The 17 sudden-event families of ReactHuman.} The families
span six motion archetypes (fall, slide \& roll, topple, articulated,
ballistic, and causal chain); each is evaluated on 18 scenes in
\S\ref{sec:experiments}, and every thumbnail is a real benchmark scene.}
\label{fig:taxonomy}
\end{figure*}

Figure~\ref{fig:taxonomy} summarizes the families. Beyond covering distinct
dynamics (free fall, friction-driven sliding, inverted-pendulum toppling,
hinge-constrained rotation, ballistic flight with restitution, momentum
transfer), families differ in \emph{decision structure}. In
\textit{object\_drop} the trajectory is trivial and the decision is dominated
by object semantics. In \textit{pendulum\_swing} and \textit{bouncing\_object}
semantics are easy but the interception point is time-varying or multimodal.
In \textit{chain\_reaction} and \textit{multi\_object} the threatening object
is initially \emph{stationary}: the model must propagate causality (A will
strike B; momentum will travel down the row) before any kinematic estimate is
possible. This factorization lets ReactHuman localize failures to perception,
causal reasoning, semantic judgment, or spatial calculation.

\subsection{Adversarial Appearance--Physics Probes}
\label{sec:adversarial}

The 82-object asset library contains 14 \emph{adversarial variants} whose
visual identity contradicts their physical parameters: a foam anvil (looks
cast-iron; light and safe to catch), a steel apple (looks like fruit; arrives
with bruising momentum), a lead-core tennis ball, a foam mirror, a hollow prop
door, a foam ceiling panel, a lead sandbag, a solid-steel can. For these
objects the appearance
prior implies the \emph{wrong} action; the correct one is recoverable only
from observed dynamics in the early frames. Section~\ref{sec:experiments} probes these objects directly: across 280
decisions no model ever questions an object's material or weight, and
disguised objects receive exactly the treatment of their genuine
look-alikes.

\subsection{Scene Generation and Ground Truth}
\label{sec:pipeline}

\begin{figure*}[t]
    \centering
    \includegraphics[width=\textwidth]{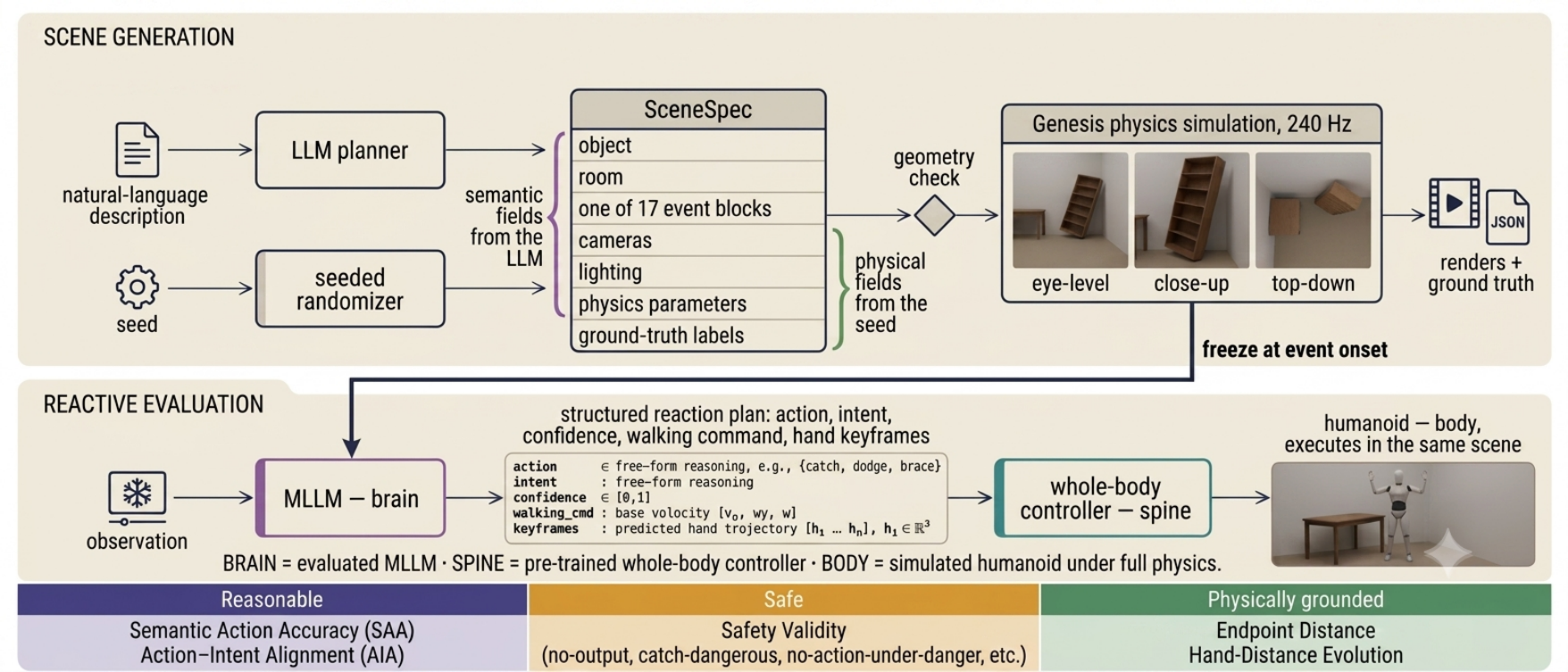}
    \caption{\textbf{The ReactHuman pipeline.} A frozen LLM planner and a seeded randomizer jointly populate a single SceneSpec. The LLM sets semantics (object, room, event family), and the seed sets all physics while Genesis simulates at 240 Hz into three synchronized views with simulator-derived ground truth. At a frozen observation, the evaluated MLLM (the brain) commits a structured reaction plan that a pre-trained controller (spine) executes on a simulated humanoid (body) in the same scene. The plan is scored on three axes: reasonable, safe, and physically grounded.}
    \label{fig:pipeline}
\end{figure*}

Every scene is produced by a three-stage pipeline (Figure~\ref{fig:pipeline}).

\paragraph{Stage 1: LLM semantic planning.}
A frozen LLM parses a natural-language description (``\emph{a cast-iron pan slides
off the counter while someone is cooking}'') into a validated selection: event
family, object (from a per-family catalogue), room type, and family-specific
hints (e.g., which intermediate surface a cascading object lands on). Outputs
are validated against the catalogue with automatic retry on malformed
responses. Objects named in descriptions but absent from the library can be
sourced automatically from Objaverse \citep{deitke2023objaverse}: candidates
retrieved by LVIS label are scored for shape plausibility, re-oriented,
welded, and registered.

\paragraph{Stage 2: Seeded physical randomization.}
A deterministic randomizer, a pure function of an integer seed, samples
every physical and visual parameter: placement and overhang, initial linear
and angular velocities, friction and restitution, room and table geometry
offsets, lighting, and camera poses (with a retry loop guaranteeing the object
remains in frame). Family-specific samplers encode each event's physics: ramp
angles are floored above the friction cone so sliding is guaranteed; ladder
lean angles bifurcate into slide-down versus tip-over regimes; chain-reaction
targets are settled to rest and only then struck by the trigger. Physics is
never scripted mid-scene; outcomes emerge from initial conditions alone.

\paragraph{Stage 3: Simulation and labeling.}
Scenes are simulated in Genesis \citep{genesis2024} with a rigid-body solver
at 240\,Hz and rendered at $1280{\times}720$/60\,fps from three cameras.
Ground truth is extracted from simulator state: the action label from the
object's catalogued properties and event kinematics; the impact point
$\mathbf{p}_{\text{impact}}\in\mathbb{R}^3$ where the trajectory crosses the
observer's reach plane; and time-to-floor from first floor contact. Each scene
ships its complete specification; re-running the specification reproduces the
simulation bit-for-bit, and geometric validators check spec--simulation
consistency before rendering.

This division of labor is deliberate: the LLM contributes semantic diversity
and natural-language grounding but never touches numbers; the seeded
randomizer owns all continuous parameters, so labels are exact and the
benchmark extends procedurally past $10^4$ scenes (with adversarial variants
injected at a configurable rate) at zero annotation cost.
\section{Metric Suite}
\label{sec:metrics}

We report five metrics per decision, each aimed at a different kind of
failure: a model can pick the right action but still be unsafe, or give a
sensible reason yet send its hands to the wrong place. Throughout, $s$
indexes scenes and $m$ models. The model never outputs an action label; we
recover the action from its motor plan (the walking command and hand
keyframes) with a fixed rule that is released with the evaluation code.

\paragraph{1. Semantic Action Accuracy (SAA).}
$\mathrm{SAA}(s,m)=\mathbf{1}\!\left[\hat a(s,m) = a^{*}(s)\right]$: whether
the chosen action matches the ground truth. This checks the action label
alone, so a model can score $\mathrm{SAA}=1$ and still break a safety rule.

\paragraph{2. Safety Validity.}
A binary flag for safety-critical mistakes, scored independently of SAA. A
decision is unsafe if it breaks any of four rules: (R1) no parseable
output; (R2) \textsc{Catch} on an object labeled dangerous; (R3)
\textsc{No\_Action} while danger is present; (R4) \textsc{Catch} or
\textsc{No\_Action} when the ground truth is \textsc{Dodge}. We record
which rule was broken. We keep safety separate from SAA on purpose:
catching a dangerous object is unsafe even when catching is a reasonable
move in general.

\paragraph{3. Physical Endpoint Distance.}
$d_{\text{end}}(s,m)=\lVert \mathbf{h}_{\text{final}}(s,m)-
\mathbf{p}_{\text{impact}}(s)\rVert_2$ in meters, the distance between the
final hand position and the true impact point. A model can choose the right
action and still put its hands in the wrong place.

\paragraph{4. Action--Intent Alignment (AIA).}
A three-level score ($0$, $0.5$, $1$) for whether the stated \texttt{intent}
matches the goal of the ground-truth action. It separates a real
misunderstanding from a right idea with a bad output: high AIA with
$\mathrm{SAA}=0$ means the model understood the scene but wrote the wrong
plan, while low AIA means it misread the scene. Intent is currently scored
by keyword matching.

\paragraph{5. Hand-Distance Evolution (HDE).}
Along the keyframe path we compute $d_i=\lVert\mathbf{h}_i-
\mathbf{p}_{\text{impact}}\rVert_2$, and report the closest the hand gets,
$d_{\min}=\min_i d_i$, together with how much it closes in, $d_1-d_{\min}$.
This shows whether the hand actually moves toward the target, which the
endpoint alone can hide: a path can wander and still finish near the impact
point.

\paragraph{Aggregation.}
We average each metric per model and per family, dropping missing values
from the denominator and reporting how many. Safety takes priority in our
plots: once a decision breaks a safety rule, that dominates its rating
regardless of the other numbers.

\section{Experiments}


\label{sec:experiments}

\subsection{Setup}

\paragraph{Models.} We evaluate seven MLLMs zero-shot through one unified API
(OpenRouter): five frontier models (Claude Opus 4.8, GPT-5.5, Gemini 2.5
Flash, Kimi K2.6, Qwen3-VL-235B) and two small open-weight models
(Qwen3-VL-30B and Gemma-3-27B), which probe whether the benchmark separates
capability tiers.\footnote{Model IDs: \url{anthropic/claude-opus-4.8},
\url{openai/gpt-5.5}, \url{google/gemini-2.5-flash},
\url{moonshotai/kimi-k2.6}, \url{qwen/qwen3-vl-235b-a22b-instruct},
\url{qwen/qwen3-vl-30b-a3b-instruct}, \url{google/gemma-3-27b-it}.}
All models receive identical prompts and output schema, observe
the same $\sim$0.6\,s multi-view window, and answer as JSON.

\paragraph{Controller.} The legs follow a pre-trained RL walking policy
\citep{rudin2022learning,unitree2024rlgym}; a scripted upper-body layer
tracks the commanded hand keyframes. Early prototypes used motion-imitation
whole-body controllers \citep{luo2023phc,tessler2024maskedmimic}, which
proved harder to steer with task-level commands and less stable under
sudden impacts, so we kept the simpler decoupled spine.

\paragraph{Scenes.} We evaluate a balanced subset of 306 scenes, 18 per
family across all 17 families, so that no family dominates the aggregate
metrics. Paired with the seven models this yields 2,138 decisions (four
lost to API errors).
\subsection{Results and Analysis}

\begin{table}[t]
\centering
\small
\setlength{\tabcolsep}{3.6pt}
\begin{tabular}{@{}lccccc@{}}
\toprule
\textbf{Model} & \textbf{SAA}\,$\uparrow$ & \textbf{Safety}\,$\uparrow$ &
\textbf{$d_{\text{end}}$(m)}\,$\downarrow$ & \textbf{AIA}\,$\uparrow$ &
\textbf{$d_{\min}$(m)}\,$\downarrow$ \\
\midrule
Claude Opus 4.8  & 50.7 & 64.7 & \textbf{1.07} & 0.41 & \textbf{1.06} \\
GPT-5.5          & \textbf{63.7} & 86.3 & 1.23 & 0.47 & 1.21 \\
Gemini 2.5 Flash & 48.7 & 78.1 & 1.28 & 0.47 & 1.28 \\
Kimi K2.6        & 40.9 & 78.2 & 1.30 & 0.32 & 1.28 \\
Qwen3-VL-235B    & 56.4 & 87.2 & 1.25 & 0.48 & 1.23 \\
\midrule
Qwen3-VL-30B     & 54.9 & 83.0 & 1.20 & 0.50 & 1.17 \\
Gemma-3-27B      & 62.4 & \textbf{88.2} & 1.26 & \textbf{0.61} & 1.25 \\
\midrule
\textbf{Mean}    & 54.0 & 80.8 & 1.23 & 0.47 & 1.21 \\
\bottomrule
\end{tabular}
\caption{Main results (\%; distances in meters) over 306 scenes spanning all
17 families (2{,}138 scene--model decisions; four lost to API errors).
Distance metrics are computed on scenes with a defined impact point; best
per column in \textbf{bold}.}
\label{tab:main}
\end{table}

\begin{figure}[t]
\centering
\includegraphics[width=0.8\columnwidth]{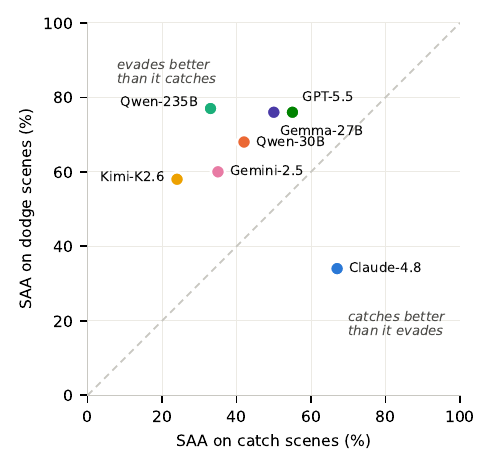}
\caption{\textbf{Action choice reflects the model, not the scene.} Each
point is one model: SAA on catch scenes (x) vs.\ dodge scenes (y); the
dashed diagonal marks equal skill. Six models sit far above it, Claude
far below, and the upper right, strong on both sides, stays empty.}
\label{fig:disposition}
\end{figure}

\paragraph{Key findings.} Table~\ref{tab:main} reports the per-model
results; three patterns stand out.
\textbf{(1) Models fail most when they need to evade.} On scenes whose
correct response is \textsc{Dodge}, 35.9\% of decisions violate a safety
rule, and this holds for every model (23--66\%). The most common error is
simply not moving: 253 of the 392 mistakes are freezing in place.
\textbf{(2) Each model has a fixed bias.} Six of seven models prefer to
retreat and score much better on dodge scenes than on catch scenes
(Figure~\ref{fig:disposition}); Claude is the opposite, catching well
(67\%) but engaging hazards it should avoid (34\% on dodge scenes). No
model adapts its behavior to the scene.
\textbf{(3) Bigger is not safer.} The 27B Gemma ranks second on accuracy
(62.4\%, just behind GPT-5.5) and first on safety (88.2\%), and both
$\sim$30B open-weight models outscore most of the frontier tier on both
columns, while the strongest flagship, Claude Opus 4.8, is the least safe
model in the table (64.7\%).

\begin{figure}[t!]
\centering
\includegraphics[width=\columnwidth]{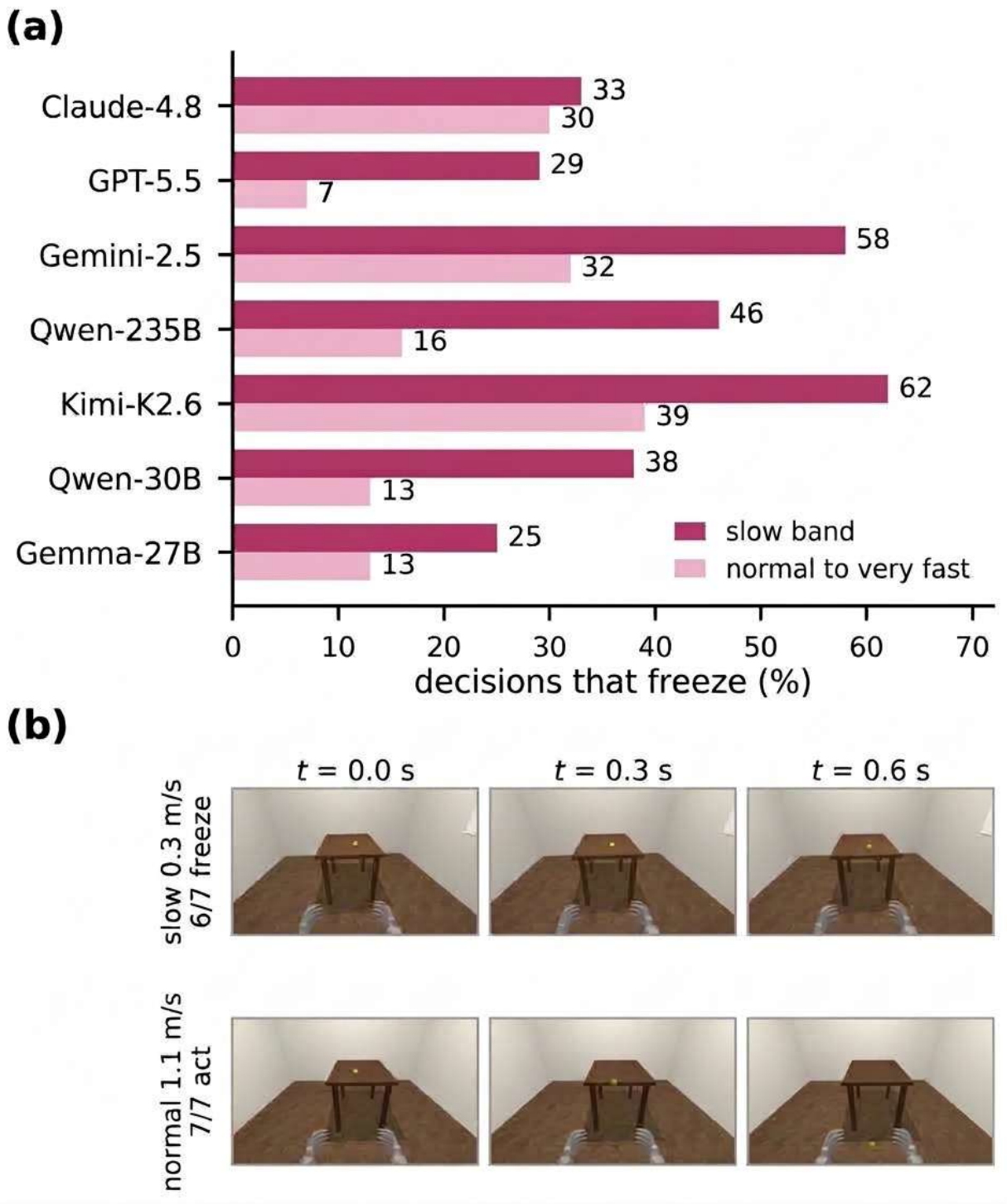}
\caption{\textbf{Slow hazards are missed by every model.} (a) Freeze rate
at the slowest speed band vs.\ the three faster bands, over the twelve
families that genuinely sweep object speed: every model freezes more on
slow hazards (pooled, 42\% vs.\ 21\%). (b) Claude's input for one
rolling-ball scene at two bands (ground truth \textsc{Catch} in both).
The slow ball still rolls toward the table edge and arrives 1\,s after
the window closes, yet six of seven models freeze, Claude included; at
the normal band none freeze and Claude catches.}

\label{fig:speed}
\end{figure}

\paragraph{Per-family breakdown.} Since six of seven models favor retreat
(Finding~2), families whose answer is \textsc{Dodge} score high for free;
we therefore compare catch and dodge families separately. Catch families
get harder as more pieces move: a single bouncing ball is caught 66.7\% of
the time, a collapsing stack of cans only 38.9\%. Dodge families get
harder as warning time shrinks: a slowly falling ceiling fixture is evaded
91.3\% of the time, an object thrown at the robot only 54.8\%, and nearly
half of the thrown-object decisions (44\%) break a safety rule.

\paragraph{Safety errors.} Of the 410 violations, freezing under danger is
the largest class (R3, 162), ahead of staying put or engaging when evasion
was required (R4, 128) and catching an object labeled dangerous (R2, 120);
R1 (unparseable output) never fired. Models fail by misjudging the hazard,
not by lacking the right action.

\paragraph{Correct but out of reach.} Saying \textsc{Catch} is not
catching. Even when models pick the right action, their hands land a
median 0.48\,m from where the object actually arrives, and 89\% of the
misses fall short. The cause is simple: models reach with the arms but do
not step forward. In half of these decisions (49\%) the robot stands more
than 0.8\,m away, farther than its arm can reach at all. A multiple-choice
benchmark would mark every one of these correct, while physical execution
shows they miss. This is why ReactHuman executes every plan.
\paragraph{Speed is perceived as binary.} Twelve of the seventeen families
sweep the object's true speed across four bands while holding everything
else fixed; in the remaining five gravity sets the pace, so we exclude
them here. On these 826 decisions, models read speed only as
moving-or-not (Figure~\ref{fig:speed}a): at the slowest band they freeze
twice as often (42\% vs.\ roughly 21\%), treating slow hazards as no
event at all even when the motion is plainly visible in their input
(Figure~\ref{fig:speed}b), and accuracy drops with them (38.1\% SAA vs.\
47--54\% at the faster bands). From normal to very fast the action
distribution barely moves. Individual decisions are also unstable: a
model keeps the same answer across all four speeds in only 27\% of scene
groups, and the slowest and fastest variants agree just 54\% of the time.
Models detect \emph{that} something moves; \emph{how fast} never enters
the decision.

\paragraph{Voting does not help.} The seven models agree unanimously on
only 16\% of scenes, and when they do, they are right 80\% of the time.
Pooling them buys little: a majority vote over all seven scores 62.1\%,
better than the average model (54.0\%) but no better than the best single
model (63.7\%). They share blind spots, and an ensemble
cannot vote shared errors away.

\paragraph{Adversarial probes.} Forty scenes plant an object whose
appearance lies about its physics: foam doors and foam ceiling panels
disguised as heavy, lead sandbags and steel cans disguised as light. The
disguise works completely: across the $40\times7=280$ decisions, no model
ever voices a doubt about an object's material or weight. For example,
the foam ceiling panels are dodged in 49 of 56 decisions, just as real
slabs would be, and the solid-steel cans get no discount for their
weight: models reach for them exactly as often as for genuinely light
cans (39\% vs.\ 39\% of decisions). For current MLLMs, seeing is
believing: physics is assigned by visual category and never revised from
observed motion.




\section{Discussion and Limitations}

\paragraph{What ReactHuman reveals.} The failures are not exotic: models
freeze in the path of hazards they plainly see, apply one fixed
disposition to every scene instead of reading it, treat speed as a binary
and slow hazards as no event, and never question an object's physics when
its appearance disagrees. None of these improve with scale, and none are
visible to an answer-only protocol. For deployment each points to a
different mitigation: safety-constrained decoding for the freeze default,
paired catch-and-dodge training to break fixed dispositions, and
motion-grounded supervision for speed and material.



\paragraph{Limitations.} Ground truth is simulator-derived; rigid-body
simulation omits deformation and shattering, and conclusions should be
validated on real hardware. The \emph{decision} is open-loop by
construction (one frozen observation, one committed plan); the humanoid
execution that follows is physically simulated but not re-planned
mid-motion. Fully closed-loop evaluation with repeated observation--action
cycles, cloth and fluid events, and native VLA policies are natural
extensions. Finally, the evaluated models are moving targets: results reflect
the specific API snapshots queried, and the standing benchmark, not any one
number, is the contribution.


\section{Conclusion}

ReactHuman reframes physical-reasoning evaluation from passive question
answering to reactive, safety-critical decision-making. Its 17 event
families, exact simulator ground truth, adversarial appearance--physics
probes, and five-metric diagnostic suite jointly measure not just whether an
embodied MLLM \emph{knows} physics, but whether it can act safely within
it, and localize the failure when it cannot. All scenes are bit-for-bit
reproducible and the LLM-planned, seed-deterministic pipeline extends the
benchmark without annotation cost. Across seven models, reactive safety
proves far from solved: models mishandle roughly one hazard in three, act
from a fixed disposition rather than the scene in front of them, and do not
improve with scale. We release the dataset to support further study.

\section*{Acknowledgments}
This work is supported by the Canada NSERC
Discovery Grant (RGPIN-2021-03115).

\bibliography{aaai2027}


\end{document}